\documentclass{article} 
\usepackage{iclr2023_conference,times}

\usepackage{amsmath,amsfonts,bm}

\def\eqref#1{equation~\ref{#1}}

\def\1{\bm{1}}

\DeclareMathAlphabet{\mathsfit}{\encodingdefault}{\sfdefault}{m}{sl}
\SetMathAlphabet{\mathsfit}{bold}{\encodingdefault}{\sfdefault}{bx}{n}

\pdfoutput=1

\usepackage[english]{babel}

\usepackage[letterpaper,top=2cm,bottom=2cm,left=3cm,right=3cm,marginparwidth=1.75cm]{geometry}

\usepackage{amsmath}
\usepackage{graphicx}
\usepackage[colorlinks=true, allcolors=blue]{hyperref}

\usepackage{hyperref}
\usepackage{url}
\usepackage{tikz}
\usepackage{xcolor}
\usepackage{tcolorbox}
\usepackage{color,xcolor}

\usepackage{amssymb}
\usepackage{amsopn}
\usepackage{graphicx}
\usepackage{multirow}
\usepackage[english]{babel}
\usepackage{bm}
\usepackage{array}
\usepackage{setspace}
\usepackage{todonotes}

\usepackage{xspace}
\usepackage{amsfonts}
\usepackage{array,multirow}
\usepackage{pgfplots}
\usepackage{tikz}
\usepackage{verbatim}
\usepackage{subfig}
\usepackage{arydshln}
\usepackage{booktabs}
\usepackage{natbib}
\usepackage{makecell}

\title{Efficient LLM Inference with KCache}
\author{Qiaozhi He, \ Zhihua Wu \\
qiaozhihe2022@outlook.com
}

\iclrfinalcopy 

\begin{document}
\maketitle

\begin{abstract}
Large Language Models(LLMs) have had a profound impact on AI applications, particularly in the domains of long-text comprehension and generation. KV Cache \citep{pope2022efficiently} technology is one of the most widely used techniques in the industry. It ensures efficient sequence generation by caching previously computed KV states. However, it also introduces significant memory overhead. We discovered that KV Cache is not necessary and proposed a novel KCache technique to alleviate the memory bottleneck issue during the LLMs inference process.  KCache can be used directly for inference without any training process, Our evaluations show that KCache improves the throughput of popular LLMs by $40\%$ with the baseline, while keeping accuracy.
\end{abstract}

\section{Introduction}
Currently, LLMs like GPT-4 \citep{openai2023gpt4}, PaLM \citep{chowdhery2022palm,anil2023palm}, LLaMA 3 \citep{llama3modelcard} dominate in numerous natural language processing, summary, code generation, question answering, etc. However, their expensive online inference cost poses significant obstacles to the deployment of LLM-based applications. With limited computational resources, how to maximize the overall system throughput as much as possible, and improving the utilization rate of the GPU cluster becomes increasingly important. LLMs inference consists of two phases: \textit{prefill} phase and \textit{decode} phase. The \textit{decode} phase generates tokens one by one, based on the result of \textit{prefill} phase and the previous step of \textit{decode} phase, which is memory bandwidth bound. So, we need to increase the batch size to improve the system throughput, but increasing the batch size will further occupy more GPU memory.

The memory usage of LLM inference mainly consists of $3$ parts: model weights, activations, and KV Cache. For Instance, For the LLaMA2-7B model, the weights occupy around $14$GB of memory at fp16 precision. When processing a batch size of $8$ and a sequence length of $32 \times 1024$, the KV cache occupies around $128$GB of memory, with the layer-wise memory sharing strategy, the activations only occupy $2$GB of memory. 
KV Cache=$2 \times bytes \times bsdl$, where $2$ represents K Cache and V Cache, $b$ represents the batch size, $s$ represents the sequence length and $d$ represents the embedding dimension and $l$ represents the number of layers. As the batch size and sequence length increase, the memory usage of the KV Cache will increase linearly. 

Some optimizations have been proposed to alleviate the KV Cache bottleneck. Quantization compression algorithms, \citep{dong2024qaq,kang2024gear,yue2024wkvquant} have been proposed to compress the KV Cache from the $bytes$ perspective. Context window compression algorithms, \citep{zhang2023h2o,xiao2024efficient,liu2023scissorhands} have been proposed to compress the KV Cache from $s$ perspective. Adaptive computation
algorithms, \citep{schuster2022confident} early exit decoding to reduce compute, which from $l$ perspective. \citep{shazeer2019fast,ainslie2023gqa} accelerates inference by improving the structure of the Multi-Head Attention (MHA). From the K Cache and V Cache perspective, although simply offloading to CPU and reloading back to GPU during inference can alleviate the pressure on GPU memory, the current Host-to-Device (H2D) and Device-to-Host (D2H) bandwidth will become the new bottleneck for inference.

\citep{zhang2023h2o,xiao2024efficient,liu2023scissorhands} have been proposed that only pivotal tokens are important during inference, which KV Cache is compressed by deleting part of them. However, considering multi-turn question-answering scenarios, deleting parts of the KV Cache directly without a fallback mechanism is a highly risky action. A more flexible approach is to retain all KV states as much as possible and dynamically select the key information for computation. This way, since all KV states are preserved, the upper bound of accuracy can be guaranteed to be high enough. Based on this idea, an obvious method is to offload all KV states to CPU memory. Another key issue is how to dynamically select which KV states are important and copy them back to HBM from CPU memory for attention calculation. As long as this partial information can maximally preserve all semantic information, the inference accuracy can approach the theoretical upper bound as much as possible, while the partial data copying can also maximize the inference performance. We propose {\bf KCache}, During the inference process, we retain the K Cache in HBM while storing the V Cache in CPU Memory. Simultaneously, we directly utilize the softmax results from the Attention computation to filter out the key information and recall the corresponding V Cache from CPU Memory for subsequent Attention calculations. Through this simple approach, leveraging the structural characteristics of Transformer models, we effectively utilize the idle CPU memory, increasing the capacity of HBM.

In this paper, we build \textit{InferenceEngine} based on KCache that efficiently reduces the memory footprint during LLM inference, which achieved $40\%$ increased throughput and keeping accuracy. The main contributions of our work include:
\begin{itemize}
    \item We propose KCache that can be used directly for inference without any training process while improving throughput by $40\%$ while maintaining accuracy.
    \item We identified the performance and accuracy challenges in offloading the VCache to CPU memory, proposed KCache to address this challenge, and validated its effectiveness through experiments on model inference.
    \item KCache is flexible and scalable, which can be applied to transformed pre-trained models.
\end{itemize}

\begin{figure}[!t]
\centering

  
\tikzset {_a467x1clu/.code = {\pgfsetadditionalshadetransform{ \pgftransformshift{\pgfpoint{0 bp } { 0 bp }  }  \pgftransformrotate{0 }  \pgftransformscale{2 }  }}}
\pgfdeclarehorizontalshading{_gul9izqgr}{150bp}{rgb(0bp)=(1,0,0);
rgb(37.5bp)=(1,0,0);
rgb(50bp)=(1,1,0);
rgb(62.5bp)=(1,0,0);
rgb(100bp)=(1,0,0)}

  
\tikzset {_ki64dnip5/.code = {\pgfsetadditionalshadetransform{ \pgftransformshift{\pgfpoint{0 bp } { 0 bp }  }  \pgftransformrotate{0 }  \pgftransformscale{2 }  }}}
\pgfdeclarehorizontalshading{_jk9d4yjh6}{150bp}{rgb(0bp)=(0,1,0);
rgb(37.5bp)=(0,1,0);
rgb(50bp)=(1,0,0);
rgb(62.5bp)=(1,1,0);
rgb(100bp)=(1,1,0)}
\tikzset{every picture/.style={line width=0.75pt}} 

\begin{tikzpicture}[x=0.75pt,y=0.75pt,yscale=-1,xscale=1]

\draw  [draw opacity=0][fill={rgb, 255:red, 248; green, 231; blue, 28 }  ,fill opacity=0.15 ] (5,4.5) -- (386.5,4.5) -- (386.5,141) -- (5,141) -- cycle ;
\draw [color={rgb, 255:red, 208; green, 2; blue, 27 }  ,draw opacity=0.5 ]   (218.2,71) -- (96.07,166.77) ;
\draw [shift={(94.5,168)}, rotate = 321.9] [color={rgb, 255:red, 208; green, 2; blue, 27 }  ,draw opacity=0.5 ][line width=0.75]    (10.93,-3.29) .. controls (6.95,-1.4) and (3.31,-0.3) .. (0,0) .. controls (3.31,0.3) and (6.95,1.4) .. (10.93,3.29)   ;
\draw [color={rgb, 255:red, 208; green, 2; blue, 27 }  ,draw opacity=0.5 ]   (19.2,71) -- (93.27,166.42) ;
\draw [shift={(94.5,168)}, rotate = 232.18] [color={rgb, 255:red, 208; green, 2; blue, 27 }  ,draw opacity=0.5 ][line width=0.75]    (10.93,-3.29) .. controls (6.95,-1.4) and (3.31,-0.3) .. (0,0) .. controls (3.31,0.3) and (6.95,1.4) .. (10.93,3.29)   ;
\draw [color={rgb, 255:red, 74; green, 144; blue, 226 }  ,draw opacity=0.5 ]   (371.3,131) -- (299.29,167.1) ;
\draw [shift={(297.5,168)}, rotate = 333.37] [color={rgb, 255:red, 74; green, 144; blue, 226 }  ,draw opacity=0.5 ][line width=0.75]    (10.93,-3.29) .. controls (6.95,-1.4) and (3.31,-0.3) .. (0,0) .. controls (3.31,0.3) and (6.95,1.4) .. (10.93,3.29)   ;
\draw [color={rgb, 255:red, 208; green, 2; blue, 27 }  ,draw opacity=0.5 ]   (21.2,131) -- (92.71,167.1) ;
\draw [shift={(94.5,168)}, rotate = 206.78] [color={rgb, 255:red, 208; green, 2; blue, 27 }  ,draw opacity=0.5 ][line width=0.75]    (10.93,-3.29) .. controls (6.95,-1.4) and (3.31,-0.3) .. (0,0) .. controls (3.31,0.3) and (6.95,1.4) .. (10.93,3.29)   ;
\draw [color={rgb, 255:red, 74; green, 144; blue, 226 }  ,draw opacity=0.5 ]   (174.3,131) -- (295.58,167.42) ;
\draw [shift={(297.5,168)}, rotate = 196.72] [color={rgb, 255:red, 74; green, 144; blue, 226 }  ,draw opacity=0.5 ][line width=0.75]    (10.93,-3.29) .. controls (6.95,-1.4) and (3.31,-0.3) .. (0,0) .. controls (3.31,0.3) and (6.95,1.4) .. (10.93,3.29)   ;
\draw [color={rgb, 255:red, 208; green, 2; blue, 27 }  ,draw opacity=0.5 ]   (218.2,131) -- (96.42,167.43) ;
\draw [shift={(94.5,168)}, rotate = 343.35] [color={rgb, 255:red, 208; green, 2; blue, 27 }  ,draw opacity=0.5 ][line width=0.75]    (10.93,-3.29) .. controls (6.95,-1.4) and (3.31,-0.3) .. (0,0) .. controls (3.31,0.3) and (6.95,1.4) .. (10.93,3.29)   ;
\draw [color={rgb, 255:red, 74; green, 144; blue, 226 }  ,draw opacity=0.5 ]   (371.3,71) -- (96.39,167.34) ;
\draw [shift={(94.5,168)}, rotate = 340.69] [color={rgb, 255:red, 74; green, 144; blue, 226 }  ,draw opacity=0.5 ][line width=0.75]    (10.93,-3.29) .. controls (6.95,-1.4) and (3.31,-0.3) .. (0,0) .. controls (3.31,0.3) and (6.95,1.4) .. (10.93,3.29)   ;
\draw [color={rgb, 255:red, 74; green, 144; blue, 226 }  ,draw opacity=0.5 ]   (172.3,71) -- (95.75,166.44) ;
\draw [shift={(94.5,168)}, rotate = 308.73] [color={rgb, 255:red, 74; green, 144; blue, 226 }  ,draw opacity=0.5 ][line width=0.75]    (10.93,-3.29) .. controls (6.95,-1.4) and (3.31,-0.3) .. (0,0) .. controls (3.31,0.3) and (6.95,1.4) .. (10.93,3.29)   ;
\draw  [color={rgb, 255:red, 208; green, 201; blue, 201 }  ,draw opacity=1 ][fill={rgb, 255:red, 225; green, 217; blue, 217 }  ,fill opacity=1 ] (210,38.2) .. controls (210,33.67) and (213.67,30) .. (218.2,30) -- (371.3,30) .. controls (375.83,30) and (379.5,33.67) .. (379.5,38.2) -- (379.5,62.8) .. controls (379.5,67.33) and (375.83,71) .. (371.3,71) -- (218.2,71) .. controls (213.67,71) and (210,67.33) .. (210,62.8) -- cycle ;

\draw  [color={rgb, 255:red, 208; green, 201; blue, 201 }  ,draw opacity=1 ][fill={rgb, 255:red, 225; green, 217; blue, 217 }  ,fill opacity=1 ] (10,38.2) .. controls (10,33.67) and (13.67,30) .. (18.2,30) -- (171.3,30) .. controls (175.83,30) and (179.5,33.67) .. (179.5,38.2) -- (179.5,62.8) .. controls (179.5,67.33) and (175.83,71) .. (171.3,71) -- (18.2,71) .. controls (13.67,71) and (10,67.33) .. (10,62.8) -- cycle ;
\draw  [color={rgb, 255:red, 208; green, 201; blue, 201 }  ,draw opacity=1 ][fill={rgb, 255:red, 225; green, 217; blue, 217 }  ,fill opacity=1 ] (10,98.2) .. controls (10,93.67) and (13.67,90) .. (18.2,90) -- (171.3,90) .. controls (175.83,90) and (179.5,93.67) .. (179.5,98.2) -- (179.5,122.8) .. controls (179.5,127.33) and (175.83,131) .. (171.3,131) -- (18.2,131) .. controls (13.67,131) and (10,127.33) .. (10,122.8) -- cycle ;

\draw  [color={rgb, 255:red, 208; green, 201; blue, 201 }  ,draw opacity=1 ][fill={rgb, 255:red, 225; green, 217; blue, 217 }  ,fill opacity=1 ] (210,98.2) .. controls (210,93.67) and (213.67,90) .. (218.2,90) -- (371.3,90) .. controls (375.83,90) and (379.5,93.67) .. (379.5,98.2) -- (379.5,122.8) .. controls (379.5,127.33) and (375.83,131) .. (371.3,131) -- (218.2,131) .. controls (213.67,131) and (210,127.33) .. (210,122.8) -- cycle ;

\draw   (134,166.26) -- (134,193.24) .. controls (134,197.25) and (115.2,200.5) .. (92,200.5) .. controls (68.8,200.5) and (50,197.25) .. (50,193.24) -- (50,166.26)(134,166.26) .. controls (134,170.27) and (115.2,173.52) .. (92,173.52) .. controls (68.8,173.52) and (50,170.27) .. (50,166.26) .. controls (50,162.25) and (68.8,159) .. (92,159) .. controls (115.2,159) and (134,162.25) .. (134,166.26) -- cycle ;

\draw   (334,166.26) -- (334,193.24) .. controls (334,197.25) and (315.2,200.5) .. (292,200.5) .. controls (268.8,200.5) and (250,197.25) .. (250,193.24) -- (250,166.26)(334,166.26) .. controls (334,170.27) and (315.2,173.52) .. (292,173.52) .. controls (268.8,173.52) and (250,170.27) .. (250,166.26) .. controls (250,162.25) and (268.8,159) .. (292,159) .. controls (315.2,159) and (334,162.25) .. (334,166.26) -- cycle ;

\draw [color={rgb, 255:red, 208; green, 2; blue, 27 }  ,draw opacity=0.5 ]   (462.76,103.63) -- (490.5,103.97) ;
\draw [shift={(492.5,104)}, rotate = 180.72] [color={rgb, 255:red, 208; green, 2; blue, 27 }  ,draw opacity=0.5 ][line width=0.75]    (10.93,-3.29) .. controls (6.95,-1.4) and (3.31,-0.3) .. (0,0) .. controls (3.31,0.3) and (6.95,1.4) .. (10.93,3.29)   ;
\draw [color={rgb, 255:red, 74; green, 144; blue, 226 }  ,draw opacity=0.5 ]   (463.5,124) -- (493.51,124.44) ;
\draw [shift={(493.51,124.44)}, rotate = 180] [color={rgb, 255:red, 74; green, 144; blue, 226 }  ,draw opacity=0.5 ][line width=0.75]    (10.93,-3.29) .. controls (6.95,-1.4) and (3.31,-0.3) .. (0,0) .. controls (3.31,0.3) and (6.95,1.4) .. (10.93,3.29)   ;

\draw  [draw opacity=0][fill={rgb, 255:red, 184; green, 233; blue, 134 }  ,fill opacity=0.2 ] (5,208.5) -- (485.5,208.5) -- (485.5,342) -- (5,342) -- cycle ;
\draw  [fill={rgb, 255:red, 248; green, 231; blue, 28 }  ,fill opacity=0.25 ] (18,242) -- (88,242) -- (88,260.5) -- (18,260.5) -- cycle ;
\draw  [fill={rgb, 255:red, 208; green, 2; blue, 27 }  ,fill opacity=0.25 ] (132.5,242) -- (132.5,312) -- (114,312) -- (114,242) -- cycle ;
\draw  [fill={rgb, 255:red, 208; green, 2; blue, 27 }  ,fill opacity=0.5 ] (151,242) -- (151,312) -- (132.5,312) -- (132.5,242) -- cycle ;
\draw  [fill={rgb, 255:red, 208; green, 2; blue, 27 }  ,fill opacity=0.5 ] (169.5,242) -- (169.5,312) -- (151,312) -- (151,242) -- cycle ;
\draw  [fill={rgb, 255:red, 208; green, 2; blue, 27 }  ,fill opacity=0.5 ] (188,242) -- (188,312) -- (169.5,312) -- (169.5,242) -- cycle ;

\path  [shading=_gul9izqgr,_a467x1clu] (222,264) -- (292,264) -- (292,282.5) -- (222,282.5) -- cycle ; 
 \draw   (222,264) -- (292,264) -- (292,282.5) -- (222,282.5) -- cycle ; 

\path  [shading=_jk9d4yjh6,_ki64dnip5] (331,264) -- (366,264) -- (366,282.5) -- (331,282.5) -- cycle ; 
 \draw   (331,264) -- (366,264) -- (366,282.5) -- (331,282.5) -- cycle ; 

\draw  [fill={rgb, 255:red, 74; green, 144; blue, 226 }  ,fill opacity=0.25 ] (416.83,240) -- (416.83,310) -- (398.33,310) -- (398.33,240) -- cycle ;
\draw  [fill={rgb, 255:red, 74; green, 144; blue, 226 }  ,fill opacity=0.5 ] (435.33,240) -- (435.33,310) -- (416.83,310) -- (416.83,240) -- cycle ;

\draw [color={rgb, 255:red, 74; green, 144; blue, 226 }  ,draw opacity=0.5 ]   (407.58,274) .. controls (295.46,254.5) and (295.43,225.49) .. (292.25,202.27) ;
\draw [shift={(292,200.5)}, rotate = 81.53] [color={rgb, 255:red, 74; green, 144; blue, 226 }  ,draw opacity=0.5 ][line width=0.75]    (10.93,-3.29) .. controls (6.95,-1.4) and (3.31,-0.3) .. (0,0) .. controls (3.31,0.3) and (6.95,1.4) .. (10.93,3.29)   ;
\draw [color={rgb, 255:red, 74; green, 144; blue, 226 }  ,draw opacity=1 ]   (292,200.5) .. controls (331.11,206.94) and (419.71,193.28) .. (425.92,271.6) ;
\draw [shift={(426.08,274)}, rotate = 266.76] [fill={rgb, 255:red, 74; green, 144; blue, 226 }  ,fill opacity=1 ][line width=0.08]  [draw opacity=0] (8.93,-4.29) -- (0,0) -- (8.93,4.29) -- cycle    ;
\draw [color={rgb, 255:red, 208; green, 2; blue, 27 }  ,draw opacity=1 ]   (97.5,201) .. controls (151.13,210.75) and (180.03,240.46) .. (161.75,273.46) ;
\draw [shift={(160.25,276)}, rotate = 302.01] [fill={rgb, 255:red, 208; green, 2; blue, 27 }  ,fill opacity=1 ][line width=0.08]  [draw opacity=0] (8.93,-4.29) -- (0,0) -- (8.93,4.29) -- cycle    ;
\draw [color={rgb, 255:red, 208; green, 2; blue, 27 }  ,draw opacity=0.5 ]   (123.25,276) .. controls (71.55,280.9) and (83.97,232.98) .. (91.54,202.35) ;
\draw [shift={(92,200.5)}, rotate = 103.82] [color={rgb, 255:red, 208; green, 2; blue, 27 }  ,draw opacity=0.5 ][line width=0.75]    (10.93,-3.29) .. controls (6.95,-1.4) and (3.31,-0.3) .. (0,0) .. controls (3.31,0.3) and (6.95,1.4) .. (10.93,3.29)   ;
\draw   (310.05,278.25) -- (310.05,275.63) -- (299.25,275.63) -- (299.25,270.38) -- (310.05,270.38) -- (310.05,267.75) -- (317.25,273) -- cycle ;
\draw [color={rgb, 255:red, 208; green, 2; blue, 27 }  ,draw opacity=1 ]   (461.7,161.63) -- (490.5,161.96) ;
\draw [shift={(493.5,162)}, rotate = 180.68] [fill={rgb, 255:red, 208; green, 2; blue, 27 }  ,fill opacity=1 ][line width=0.08]  [draw opacity=0] (8.93,-4.29) -- (0,0) -- (8.93,4.29) -- cycle    ;
\draw [color={rgb, 255:red, 74; green, 144; blue, 226 }  ,draw opacity=1 ]   (461.85,178.63) -- (490.5,178.96) ;
\draw [shift={(493.5,179)}, rotate = 180.68] [fill={rgb, 255:red, 74; green, 144; blue, 226 }  ,fill opacity=1 ][line width=0.08]  [draw opacity=0] (8.93,-4.29) -- (0,0) -- (8.93,4.29) -- cycle    ;

\draw (124,221.25) node   [align=left] {\begin{minipage}[lt]{161.84pt}\setlength\topsep{0pt}
KCache Attention in Decode Stage
\end{minipage}};
\draw (96,15.25) node   [align=left] {\begin{minipage}[lt]{115.6pt}\setlength\topsep{0pt}
KCache \ in Prefill Stage
\end{minipage}};
\draw (16,318.4) node [anchor=north west][inner sep=0.75pt]    {$Q\in R^{1\times H}$};
\draw (111,318.4) node [anchor=north west][inner sep=0.75pt]    {$K^{\top } \in R^{H\times S}$};
\draw (222,318.4) node [anchor=north west][inner sep=0.75pt]    {$S\in R^{1\times S}$};
\draw (320,318.4) node [anchor=north west][inner sep=0.75pt]    {$S\in R^{1\times N}$};
\draw (394,318.4) node [anchor=north west][inner sep=0.75pt]    {$V\in R^{N\times H}$};
\draw (202.23,274.58) node  [font=\Large] [align=left] {{\Large =}};
\draw (101.08,250.2) node  [font=\Large]  {$\times $};
\draw (379.08,270.2) node  [font=\Large]  {$\times $};
\draw (189,97.2) node [anchor=north west][inner sep=0.75pt]   [align=left] {...};
\draw (27,39.4) node [anchor=north west][inner sep=0.75pt]    {$Transformer\ Layer_{0}$};
\draw (188,39.5) node [anchor=north west][inner sep=0.75pt]   [align=left] {...};
\draw (224,100.9) node [anchor=north west][inner sep=0.75pt]    {$Transformer\ Layer_{n}$};
\draw (227,41.9) node [anchor=north west][inner sep=0.75pt]    {$Transformer\ Layer_{i}$};
\draw (24,101.9) node [anchor=north west][inner sep=0.75pt]    {$Transformer\ Layer_{i+1}$};
\draw (96,185.52) node   [align=left] {\begin{minipage}[lt]{25.85pt}\setlength\topsep{0pt}
{\footnotesize HBM}
\end{minipage}};
\draw (296.5,185.52) node   [align=left] {\begin{minipage}[lt]{57.8pt}\setlength\topsep{0pt}
{\footnotesize CPU Memory}
\end{minipage}};
\draw (462.23,275.58) node  [font=\Large] [align=left] {{\Large =}};
\draw (424.78,160) node   [align=left] {\begin{minipage}[lt]{50.22pt}\setlength\topsep{0pt}
{\fontfamily{ptm}\selectfont {\footnotesize KCache Pull}}
\end{minipage}};
\draw (424.95,172.81) node   [align=left] {\begin{minipage}[lt]{49.98pt}\setlength\topsep{0pt}
{\fontfamily{ptm}\selectfont {\footnotesize VCache Pull}}
\end{minipage}};
\draw (426,101) node   [align=left] {\begin{minipage}[lt]{51.8pt}\setlength\topsep{0pt}
{\fontfamily{ptm}\selectfont {\footnotesize KCache Push}}
\end{minipage}};
\draw (427.3,120.56) node   [align=left] {\begin{minipage}[lt]{52.75pt}\setlength\topsep{0pt}
{\fontfamily{ptm}\selectfont {\footnotesize VCache push}}
\end{minipage}};

\end{tikzpicture}

\caption
{
{\bf Illustration of the KCache.}
{
During $prefill$ phase, the computation results of each layer $push$ to the HBM. After that, the part of V Cache will be copied to the CPU asynchronously, while releasing the GPU memory occupied by this part of the V Cache. During $decode$ phase, $K$ states will be pushed and pulled as KV Cache. However, we will calculate the $topN$ of attention scores, and based on the indices of the topN results, we will pull the corresponding V Cache from the CPU to the HBM in real-time to complete the subsequent computation.
}
}
\label{Illustration_of_the_Kcache}
\end{figure}
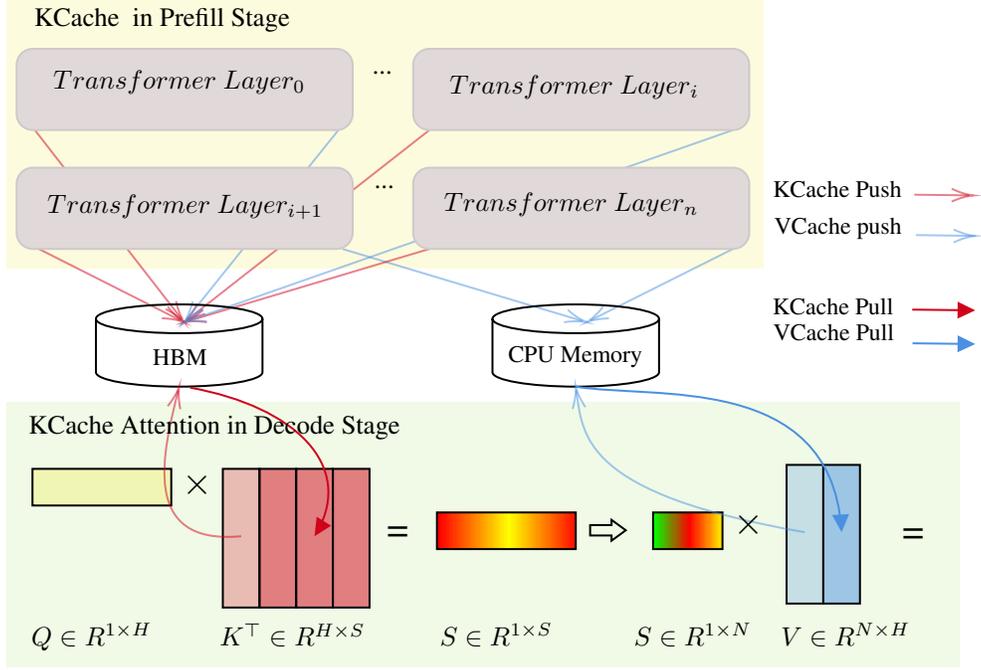


\section{Background}
In this section, we present some basic knowledge of LLMs, which include \textit{autoregressive inference}, \textit{prefill} and \textit{decode}.

LLMs are essentially based on a decoder-only architecture, which consists of $L$ stacked blocks. Each block includes two modules: a multi-head attention (MHA) \citep{vaswani2023attention} and a fully connected feed-forward network (FFN). An input tensor $x$ $\in$ ${R}^{b \times s \times d}$, where $b$ represents the batch size, $s$ represents the sequence length of input, and $d$ represents the hidden dimension of the model. MHA maps the input $x$ to different subspaces using $n$ heads: $H^i=softmax(Q^i(K^i)^{\top}/\sqrt{d_h})V^i$, $MHA(x)=Concat(H^1,H^2,...,H^{n-1},H^n)W_o$, where $Q^i=xW_{q_i}$, $K^i=xW_{k_i}$, $V^i=xW_{v_i}$, and $W_{q_i} \in R^{h \times h}$, $W_{k_i} \in R^{h \times h}$, $W_{v_i} \in R^{h \times h}$ are trainable weights, $h$ represents the hidden dimension of per head. $d_h=d/n$. FFN take SwiGLU \citep{shazeer2020glu} for examle, $FFN_{SwiGLU}(x, W, V, W_2) = (\sigma(xW) \bigotimes xV )W_2$, where $W \in R^{d \times 8/3d}$, $V \in R^{d \times 8/3d}$, $W_2 \in R^{8/3d \times d}$, $\sigma$ is unit of activation. LLMs process a sequence of words named \textit{prompt} and generate some new words. The autoregressive inference means that the token generated at the current moment depends on the token generated at the previous moment. The process of handling user prompts is called \textit{prefill}, and it only needs to be done once. The process of generating all output tokens one by one in autoregression is called \textit{decode} and needs to be executed continuously.
During the \textit{prefill} phase, taking prompt as input and computation in parallel using matrix-matrix multiplications. During the \textit{decode} phase, which performs the same operations as \textit{prefill}, but only takes one token as input and computation with KV Cache using vector-matrix multiplications.




\section{Method}

\subsection{KCache}
\begin{table*}[!ht]
\centering
\small
\begin{spacing}{1.1}
\setlength{\tabcolsep}{1.3mm}{
\begin{tabular}{lccccccccc}
\toprule
{\bf } & {\bf Submodule} & {\bf FLOPs} & {\bf I/O(byte)} & {\bf Arithmetic Intensity} & {\bf }\\
\midrule
\multirow{4}{*}{MHA} & $ Q=xW_{q}$, $K=xW_{k}$, $V=xW_{v}$ & $6bd^2$ & $12bd + 6d^2$ & $\frac{1}{\frac{2}{d}+\frac{1}{b}} \approx b$ \\
& $ S=softmax(QK^{\top}/\sqrt{d_h})$ & $2bsd$ & $2bnh+2bnhs+2bns$ & $\frac{1}{1+\frac{1}{h}+\frac{1}{s}} \approx 1$ \\
& $ A=SV$ & $2bsd$ & $2bns + 2bsd + 2bd$ & $\frac{1}{1+\frac{1}{h}+\frac{1}{s}} \approx 1$ \\
& $ O=AW_o$ & $2bd^2$ & $4bd + 2d^2$ & $\frac{1}{\frac{2}{d}+\frac{1}{b}} \approx b$ \\
\midrule
\multirow{4}{*}{KCache MHA} & $ Q=xW_{q}$, $K=xW_{k}$, $V=xW_{v}$ & $6bd^2$ & $12bd + 6d^2$ & $\frac{1}{\frac{2}{d}+\frac{1}{b}} \approx b$ \\
& $ \tilde{S}=TopN(softmax(QK^{\top}/\sqrt{d_h}))$ & $2bsd$ & $2bnh+2bnhs+2bnN$ & $\frac{1}{1+\frac{N}{sh}+\frac{1}{s}} \approx 1$ \\
& $ \tilde{A}=\tilde{S}Part(V)$ & $2bNd$ & $2bnN + 2bNd + 2bd$ & $\frac{1}{1+\frac{1}{h}+\frac{1}{s}} \approx 1$ \\
& $ \tilde{O}=\tilde{A}W_o$ & $2bd^2$ & $4bd + 2d^2$ & $\frac{1}{\frac{2}{d}+\frac{1}{b}} \approx b$ \\
\bottomrule 
\end{tabular}}
\end{spacing}
\caption{
\label{tab1}
{\bf MHA FLOPs and I/O(byte) in \textit{decode} phase.}
$N$ denotes that the value of $N$ selected for the TopN operation. 
}
\end{table*}


In long-context scenarios, users typically ask multiple rounds of questions based on a long sequence, with each question potentially focusing on different segments of the long context. To maximize the accuracy of results in each round, we avoid reducing or compressing the KV states, thus ensuring the upper bound of model effectiveness. However, simply offloading KV states to CPU memory and reloading them to the GPU during inference would significantly increase the end-to-end inference time. Therefore, to balance model effectiveness and inference latency, we must find a way to reload only the necessary information back to HBM, which implies the need for a module to determine which information is important. Fortunately, considering the meaning of the Key and Value pairs in the Attention mechanism, where Key is used to compute the relevance with Query and Value represents the actual information associated with Key, it inspires us to offload a portion of the K Cache and V Cache to CPU memory.

Figure \ref{Illustration_of_the_Kcache} shows the method of KCache. We keep K Cache and first of $0...i$ layers V Cache in HBM and keep other V Cache in CPU memory. During computation, The attention computation is adjusted from $softmax(QK^{\top}/\sqrt{d_h})$ to $TopN(softmax(QK^{\top}/\sqrt{d_h}))$. Since the K Cache is still stored in HBM, the computation of $QK^{\top}$ is not affected. After the softmax operation, TopN selects the $N$ most relevant results. We dynamically and flexibly move the corresponding vectors of the V Cache to HBM in real-time based on the attention scores, to participate in subsequent calculations.

Based on the proposed KCache method, intuitively, as $N$ increases, the model's inference accuracy will approach that of the full KV Cache, but it will also increase the data copying overhead, leading to performance degradation. Whether there exists a perfect balance between inference performance and inference accuracy requires quantitative analysis. In the following sections, we provide an analysis from both the accuracy and performance perspectives.




\subsection{Analysis of KCache Performance }
\label{sec:Performance}

During the $prefill$ phase, the part of V Cache needs to be asynchronously copied to the CPU memory. We hope that the computation time for each layer can overlap the data copying time of the previous layer. There are $2bsd$ bytes data needed to transmit from Device to Host for each transformer block, and $22bsd^2 + 4bs^2d$ floating point operations(FLOPs) for each Transformer Block. Let:
\begin{equation}
\frac{22bsd^2 + 4bs^2d}{FLOPS} > \frac{2bsd}{Bandwidth_{D2H}}
\end{equation}
\begin{equation}
11d+2s>\frac{FLOPs}{Bandwidth_{D2H}}
\end{equation}
Take NVIDIA A100 (80GB) GPU for instance, $d=4096$ for LLaMA2-7B, which means computation will overlap the transmission.
During the $decode$ phase, the Multi-Head Attention (MHA) module is a typical memory-bound task, as evidenced by its Arithmetic Intensity shown in Table \ref{tab1}. The Arithmetic Intensity is defined as the ratio of floating-point operations (FLOPs) to I/O bytes. This indicates that the computation time of the MHA module during decoding is strongly dependent on the amount of memory access. Notably, the performance of the MHA module in the $decode$ phase is independent of the hidden size and sequence length and is solely influenced by the batch size. This observation leads to an expectation: the computation time and data transfer time of the proposed KCache MHA module can be less than the conventional KV cache MHA implementation. Let:
\begin{equation}
\frac{bnNh}{Bandwidth_{H2D}} < \frac{(2bns + 2bsd + 2bd) - (2bnN + 2bNd + 2bd)}{Bandwidth_{GPU}}
\end{equation}
\begin{equation}
\frac{s}{N} > \frac{Bandwidth_{GPU}}{Bandwidth_{H2D}}
\end{equation}
Take NVIDIA A$100$($80$GB) GPU ($2039$GB/s GPU Memory Bandwidth and $32$GB/s H2D Bandwidth) for instance, which means KCache performance will not decrease when $s/N>64$.

\subsection{Analysis of KCache Accuracy}
During the \textit{prefill} phase, the \textit{Value} tensors are asynchronously offloaded to CPU memory, which does not affect the inference accuracy and performance. During the \textit{decode} phase, it is necessary to reduce the amount of data transferred from host to device. Based on $S_{b,i}=softmax(Q^{b,i}(K^{b,i})^{\top}/\sqrt{d_h}) \in {R}^{1 \times s}$, $A_{b,i}=S_{b,i}V_{b,i} \in {R}^{1 \times d}$, where $b$ represents one instance of batch and $i$ represents one of head. If the result of $S_{b,i}$ is sparse enough, the impact of the corresponding value of $A_{b,i}$ on the final result will be negligible. In \ref{sec:Experiments}, The accuracy of KCache will be further verified.

\begin{table*}[!ht]
\centering
\footnotesize
\begin{spacing}{1.5}
\setlength{\tabcolsep}{1.3mm}{
\begin{tabular}{lccccccccccccccc}
\toprule
& \multicolumn{3}{c}{LLaMA2-7B} & \multicolumn{3}{c}{LLaMA2-13B} & \multicolumn{3}{c}{LLaMA3-8B} & \multicolumn{3}{c}{Mistral-13B} \\
\midrule
{\bf Config} & {\makecell[c]{\bf BBH \\ 3-shot}} & {\makecell[c]{\bf GSM \\ 8-shot}} & {\makecell[c]{\bf TriQA \\ 5-shot}} & {\makecell[c]{\bf BBH \\ 3-shot}} & {\makecell[c]{\bf GSM \\ 8-shot}} & {\makecell[c]{\bf TriQA \\ 5-shot}} & {\makecell[c]{\bf BBH \\ 3-shot}} & {\makecell[c]{\bf GSM \\ 8-shot}} & {\makecell[c]{\bf TriQA \\ 5-shot}} & {\makecell[c]{\bf BBH \\ 3-shot}} & {\makecell[c]{\bf GSM \\ 8-shot}} & {\makecell[c]{\bf TriQA \\ 5-shot}} \\ 
\midrule
Baseline & {\bf 39.87} & 15.09 & 64.24 & {\bf 47.69} & 25.78 & 70.64 & 62.39 & 47.61 & 71.71 & 56.21 & {\bf 40.33} & {\bf 70.94}\\
\midrule
$N=32$ $L=0$ & 33.19 & 5.46 & 63.57 & 40.73 & 10.84 & {\bf 70.67} & 56.46 & 46.7 & 71.67 & 46.54 & 34.5 & 70.84\\
$N=32$ $L=1$ & 36.94 & 12.59 & 64.16 & 45.97 & 25.32 & 70.53 & 56.38 & 46.32 & 71.72 & 52.08 & 36.62 & 70.85\\
$N=32$ $L=2$ & 37.08 & 14.25 & 64.12 & 46.17 & 26.31 & 70.49 & 56.43 & 46.17 & 71.7 & 52.17 & 37.38 & 70.85\\
$N=32$ $L=3$ & 36.63 & 14.33 & 64.14 & 45.68 & 25.93 & 70.49 & 56.41 & 45.87 & 71.69 & 52.37 & 36.85 & 70.82\\
\midrule
$N=64$ $L=0$ & 35.75 & 6.14 & 62.48 & 45.4 & 14.4 & 69.73 & 60.99 & 47.54 & 71.67 & 54.19 & 38.29 & 70.87\\
$N=64$ $L=1$ & 38.03 & 13.42 & 64.18 & 46.52 & 26.84 & 70.61 & 61.23 & 48.67 & 71.66 & 54.52 & 37.15 & 70.88\\
$N=64$ $L=2$ & 37.97 & {\bf 15.31} & 64.18 & 46.98 & {\bf 27.22} & 70.61 & 61.1 & 47.99 & 71.68 & 54.38 & 38.44 & 70.88\\
$N=64$ $L=3$ & 38.07 & 15.16 & 64.22 & 46.89 & 25.55 & 70.59 & 61.1 & 47.84 & {\bf 71.73} & 54.72 & 39.04 & 70.88\\
\midrule
$N=128$ $L=0$ & 37.26 & 8.49 & 64.07 & 45.54 & 18.12 & 70.63 & 62.69 & 47.84 & 71.72 & 56.18 & 37.83 & {\bf 70.94}\\
$N=128$ $L=1$ & 38.57 & 14.25 & 64.24 & 47.23 & 25.78 & 70.66 & {\bf 62.79} & 48.07 & 71.7 & 56 & 37.91 & 70.93\\
$N=128$ $L=2$ & 39 & 15.01 & 64.24 & 47.18 & 25.17 & {\bf 70.67} & {\bf 62.79} & {\bf 48.98} & 71.72 & 55.89 & 37.98 & 70.92 \\
$N=128$ $L=3$ & 38.89 & 15.09 & {\bf 64.25} & 47.17 & 25.55 & {\bf 70.67} & 62.52 & 48.52 & 71.72 & {\bf 56.26} & 38.44 & 70.92\\
\bottomrule 
\end{tabular}}
\end{spacing}
\caption{
\label{tab2}
KCache results for LLaMA2-7B, LLaMA2-13B, LLaMA3-8B and Mistral-13B. GSM denotes GSM8K, and TriQA denotes TriviaQA. BBH, GSM8K and TriviaQA are measured in accuracy. $N$ denotes that the value of $N$ selected for the TopN operation. $L$ denotes that the layer of VCache allocated on HBM, which means VCache of $layer_0$ and $layer_1$ was allocated on HBM and other layers on CPU memory when $L=2$. We provide the score of LLaMA3-8B with KCache where $N=128$ and $L=1$ on each subject of BBH in Table \ref{tab3}.
}
\end{table*}

\section{Experiments}
\label{sec:Experiments}

\subsection{Setup}

{\bf Models and Datasets.} All models are based on decoder-only transformers, We evaluate KCache on four open-source LLMs: LLaMA2-7B, LLaMA2-13B\citep{touvron2023llama}, LLaMA3-8B \citep{llama3modelcard} and Mistral-7B\citep{jiang2023mistral}. We choose 3 benchmarks: BBH, GSM8K and TriviaQA. BBH \citep{suzgun2022challenging} is a suite of 23 challenging BIG-Bench tasks. GSM8K \citep{cobbe2021training}, a dataset of 8.5k high-quality linguistically diverse grade school math word problems. TriviaQA \citep{joshi2017triviaqa}, a reading comprehension dataset containing over 650K question-answer-evidence triples.

\subsection{Results} 
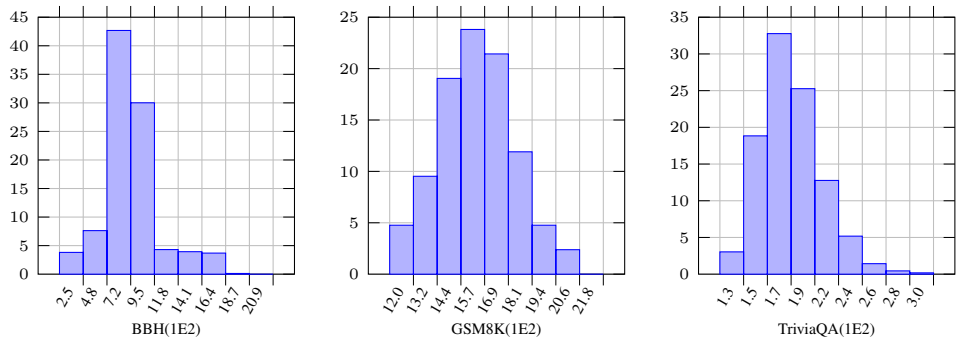
\begin{figure*}[!t]
  \centering
  \begin{tikzpicture}
    \begin{axis}[
        ybar interval,
        width=0.32\textwidth,
        height=5cm,
        bar width=5pt,
        ymin=0, ymax=45,
        xlabel=BBH(1E2),
        xticklabels={2.5, 4.8, 7.2, 9.5, 11.8, 14.1, 16.4, 18.7, 20.9, 23.3, 25.6},
        xtick={1,2,3,4,5,6,7,8,9,10},
        ytick={0,5,10,15,20,25,30,35,40,45},
        ymajorgrids,
        font=\tiny,
        xticklabel style={rotate=60, anchor=east, inner sep=1pt} 
    ]
    \addplot[fill=blue!30, draw=blue] coordinates {(1,3.813)(2,7.626)(3,42.681)(4,30.012)(5,4.305)(6,3.936)(7,3.69)(8,0.123)(9,0.0)(10,3.813)};
    \end{axis}
  \end{tikzpicture}
  \hspace{0.3cm}
  \begin{tikzpicture}
    \begin{axis}[
        ybar interval,
        width=0.32\textwidth,
        height=5cm,
        bar width=5pt,
        ymin=0, ymax=25,
        xlabel=GSM8K(1E2),
        xticklabels={12.0, 13.2, 14.4, 15.7, 16.9, 18.1, 19.4, 20.6, 21.8, 23.0, 24.3},
        xtick={1,2,3,4,5,6,7,8,9,10},
        ytick={0,5,10,15,20,25},
        ymajorgrids,
        font=\tiny,
        xticklabel style={rotate=60, anchor=east, inner sep=1pt} 
    ]
    \addplot[fill=blue!30, draw=blue] coordinates {(1,4.762)(2,9.524)(3,19.048)(4,23.81)(5,21.429)(6,11.905)(7,4.762)(8,2.381)(9,0)(10,2.381)};
    \end{axis}
  \end{tikzpicture}
  \hspace{0.3cm}
  \begin{tikzpicture}
    \begin{axis}[
        ybar interval,
        width=0.32\textwidth,
        height=5cm,
        bar width=5pt,
        ymin=0, ymax=35,
        xlabel=TriviaQA(1E2),
        xticklabels={1.3, 1.5, 1.7, 1.9, 2.2, 2.4, 2.6, 2.8, 3.0, 3.2, 3.4},
        xtick={1,2,3,4,5,6,7,8,9,10},
        ytick={0,5,10,15,20,25,30,35},
        ymajorgrids,
        font=\tiny,
        xticklabel style={rotate=60, anchor=east, inner sep=1pt} 
    ]
    \addplot[fill=blue!30, draw=blue] coordinates {(1,3.036)(2,18.839)(3,32.768)(4,25.268)(5,12.768)(6,5.179)(7,1.429)(8,0.446)(9,0.179)(10,0.089)};
    \end{axis}
  \end{tikzpicture}
  \caption{Prompt length of BBH, GSM8K and TriviaQA.}\label{fig:data_length}
\end{figure*}

\begin{table*}[!ht]
\centering
\footnotesize
\begin{spacing}{1.5}
\setlength{\tabcolsep}{1.3mm}
\begin{tabular}{lccccccccccccccc}
\toprule
\multicolumn{2}{c}{\bfseries Configuration} & {\bfseries KVCache} & \multicolumn{3}{c}{\bfseries KCache} \\ 
\midrule
{\makecell[c]{\bfseries Input \\ \bfseries Output}} & {\bfseries Batch Size} & {\makecell[c]{\bfseries Throughput \\ tokens/second}} & {\makecell[c]{\bfseries Throughput \\ $N=32$}} & {\makecell[c]{\bfseries Throughput \\ $N=64$}} & {\makecell[c]{\bfseries Throughput \\ $N=128$}}\\
\midrule
\multirow{3}{*}{\centering $\boldsymbol{1k/1k}$}
& 1  & 55.3 & 43.7 & 43.4 & 42.9 \\

& 8  & 321.0 & 277.1 & 256.2 & 222.2 \\

& 16 & {\bf 485.9} & 441.0 & 389.7 & 315.7 \\

\midrule

\multirow{3}{*}{\centering $\boldsymbol{4k/1k}$}
& 1  & 50.7 & 41.9 & 41.7 & 41.2 \\

& 8  & 212.0 & 225.9 & 211.8 & 188.0 \\

& 14 & 251.2 & 290.8 & 267.8 & 231.4 \\

& 23 & OOM & {\bf 349.2} & 316.2 & 266.2 \\
\midrule

\multirow{3}{*}{\centering $\boldsymbol{7k/1k}$} 
& 1  & 46.7 & 40.5 & 40.1 & 39.6 \\

& 8  & 158.4 & 189.8 & 180.1 & 162.6 \\

& 13 & OOM & {\bf 223.2} & 209.5 & 186.1 \\
\midrule

\multirow{3}{*}{\centering $\boldsymbol{15k/1k}$} 
& 1  & 38.3 & 36.5 & 36.4 & 36.0 \\

& 3  & 65.9 & 76.6 & 75.8 & 73.7 \\

& 5 & OOM & {\bf 102.5} & 100.4 & 95.1 \\
\bottomrule 
\end{tabular}
\end{spacing}
\caption{KCache throughput on LLaMA2-7B. OOM denotes Out of Memory Error. KCache demonstrated performance advantages when handling contexts longer than 4K, and this advantage further increased as the context length grew. When reaching 15K, KCache exhibited over $40\%$ higher throughput compared to the baseline.}
\label{tab4}
\end{table*}
{\bf Accuracy.} We run all tasks with \citep{eval-harness} for fair comparison. Table \ref{tab2} shows experimental results about the Few-shot performance. Fig \ref{fig:data_length} shows prompt length on different datasets. 
\begin{itemize}
    \item KCache essentially maintained accuracy without loss, and even achieved better performance across multiple datasets and models.
    \item $L$ has a relatively small impact on the model accuracy, especially when the model performance is sufficiently strong. If the model performance is relatively weak, it is recommended to set a larger $L$.
    \item A larger $N$ would achieve higher accuracy. When $N=128$, KCache maintained the same accuracy as the baseline, or even higher. We believe that TopN regularized the softmax and further filtered out noise information.
    \item Our experiments on three datasets validated that for context lengths around 2K or less, setting N to either 64 or 128 did not significantly impact the accuracy.
\end{itemize}



{\bf Performance.} We further evaluated the end-to-end performance of KCache in our InferenceEngine and conducted experiments on GPU, which has 64GB memory with 1TB GPU memory bandwidth and 180TFLOPS. We evaluate LLaMA2-7B. Table \ref{tab4} shows the experimental result. Overall, The experimental results further validate the analysis in \ref{sec:Performance}, where KCache demonstrates performance advantages when $S >> N$. Simultaneously, based on the results in \ref{tab2}, KCache achieved a $40\%+$ throughput improvement in inference with 15K context length with $N=128$.


\section{Conclusion}

In this work, We propose KCache, an efficient inference technique for large language models. Particularly in long-context inference scenarios, KCache demonstrates a $40\%+$ throughput improvement. This approach does not require any training and applies to various mainstream structures such as MHA and GQA. In the future, we will further explore strategies based on KCache.

\bibliographystyle{main}
\bibliography{main}

\appendix
\section{Appendix}



\begin{table*}[!ht]
\centering
\small
\begin{spacing}{1.1}
\setlength{\tabcolsep}{1.3mm}{
\begin{tabular}{lccccccccc}
\toprule
{\bf BBH} & {\bf KVCache} & {\bf KCache} \\
\midrule
bbh\_cot\_fewshot\_boolean\_expressions & 87.60 & 88.40 \\
bbh\_cot\_fewshot\_causal\_judgement & 47.59 & 54.55 \\
bbh\_cot\_fewshot\_date\_understanding & 82.40 & 82.40 \\
bbh\_cot\_fewshot\_disambiguation\_qa & 46.00 & 60.80 \\
bbh\_cot\_fewshot\_dyck\_languages & 10.00 & 10.80 \\
bbh\_cot\_fewshot\_formal\_fallacies & 53.60 & 53.60 \\
bbh\_cot\_fewshot\_geometric\_shapes & 41.20 & 41.20 \\
bbh\_cot\_fewshot\_hyperbaton & 92.80 & 92.00 \\
bbh\_cot\_fewshot\_logical\_deduction\_five\_objects & 44.40 & 44.80 \\
bbh\_cot\_fewshot\_logical\_deduction\_seven\_objects & 36.40 & 34.40 \\
bbh\_cot\_fewshot\_logical\_deduction\_three\_objects & 79.20 & 76.40 \\
bbh\_cot\_fewshot\_movie\_recommendation & 88.40 & 89.60 \\
bbh\_cot\_fewshot\_multistep\_arithmetic\_two & 31.60 & 37.20 \\
bbh\_cot\_fewshot\_navigate & 88.80 & 86.80 \\
bbh\_cot\_fewshot\_object\_counting & 82.80 & 82.40 \\
bbh\_cot\_fewshot\_penguins\_in\_a\_table & 69.18 & 67.81 \\
bbh\_cot\_fewshot\_reasoning\_about\_colored\_objects & 76.40 & 74.00 \\
bbh\_cot\_fewshot\_ruin\_names & 69.20 & 69.60 \\
bbh\_cot\_fewshot\_salient\_translation\_error\_detection & 54.00 & 52.00 \\
bbh\_cot\_fewshot\_snarks & 60.11 & 67.98 \\
bbh\_cot\_fewshot\_sports\_understanding & 96.00 & 96.00 \\
bbh\_cot\_fewshot\_temporal\_sequences & 71.20 & 66.80 \\
bbh\_cot\_fewshot\_tracking\_shuffled\_objects\_five\_objects & 36.80 & 40.40 \\
bbh\_cot\_fewshot\_tracking\_shuffled\_objects\_seven\_objects & 32.00 & 28.80 \\
bbh\_cot\_fewshot\_tracking\_shuffled\_objects\_three\_objects & 61.60 & 59.60 \\
bbh\_cot\_fewshot\_web\_of\_lies & 100.00 & 100.00 \\
bbh\_cot\_fewshot\_word\_sorting & 43.60 & 38.40 \\
\bottomrule 
\end{tabular}}
\end{spacing}
\caption{
\label{tab3}
{\bf The scores of each subject in BBH of LLaMA3-8B.}
}
\end{table*}

\end{document}